\documentclass[conference, twocolumn]{IEEEtran}

\usepackage{graphicx}
\usepackage{array}
\usepackage{breqn}

\usepackage[utf8]{inputenc}
\usepackage[T1]{fontenc}

\usepackage{listings}
\usepackage{subcaption}
\usepackage{natbib}
\usepackage[table,xcdraw]{xcolor}
\usepackage{amsmath}
\usepackage{bbm}
\usepackage{pifont}
\usepackage{amssymb}
\definecolor{mydarkblue}{rgb}{0,0.08,0.65}
\usepackage[colorlinks=true,
    linkcolor=mydarkblue,
    citecolor=mydarkblue,
    filecolor=mydarkblue,
    urlcolor=mydarkblue]{hyperref}
\usepackage{cleveref}
\usepackage{soul}
\usepackage[shortlabels]{enumitem}
\usepackage{url}
\usepackage{color}
\usepackage{tikz}
\usepackage{balance}
\usepackage{multirow}
\usepackage{comment}
\usepackage{wrapfig,booktabs}

\usepackage[symbol]{footmisc}
\usepackage{tablefootnote}
\usepackage{tabularx}
\usepackage{placeins}
\usepackage{algorithm}
\usepackage{algpseudocode}
\usepackage{ragged2e}
\usepackage{cuted}
\usepackage{float}
\usepackage{caption}

\usepackage{multicol}

\usepackage{dblfloatfix}

\definecolor{codegreen}{rgb}{0,0.6,0}
\definecolor{codegray}{rgb}{0.5,0.5,0.5}
\definecolor{codepurple}{rgb}{0.58,0,0.82}
\definecolor{backcolour}{rgb}{0.95,0.95,0.92}

\makeatletter
\def\blfootnote{\xdef\@thefnmark{}\@footnotetext}
\makeatother

\lstdefinestyle{mystyle}{
  backgroundcolor=\color{backcolour},   commentstyle=\color{codegreen},
  keywordstyle=\color{magenta},
  numberstyle=\tiny\color{codegray},
  stringstyle=\color{codepurple},
  basicstyle=\ttfamily\footnotesize,
  breakatwhitespace=false,
  breaklines=true,
  captionpos=b,
  keepspaces=true,
  numbers=left,
  numbersep=5pt,
  showspaces=false,
  showstringspaces=false,
  showtabs=false,
  tabsize=2,
}

\AtBeginDocument{%
  \providecommand\BibTeX{{%
    \normalfont B\kern-0.5em{\scshape i\kern-0.25em b}\kern-0.8em\TeX}}}

\IEEEoverridecommandlockouts

\begin{document}

\title{Equilibrium Propagation and Hamiltonian Inference in the Diffusive Fitzhugh-Nagumo Model}

\newcommand{\corr}{\textsuperscript{*}}

\author{Jack Kendall\corr \\
        \thanks{Work performed primarily while at Rain Neuromorphics}
\\[0.5em]
\textbf{Zyphra}
\\
San Francisco, CA \\
\IEEEauthorblockA{\textsuperscript{*}Correspondence: \texttt{jack@zyphra.com}, \texttt{tom@zyphra.com}}
}

\maketitle

\setcounter{page}{1}

\begin{abstract}\normalfont\mdseries
    In this work, we extend the Equilibrium Propagation framework to skew-gradient systems and show an equivalence between deep Energy-Based Models and Hamiltonian neural networks. We focus on networks of diffusively coupled Fitzhugh-Nagumo neurons as a prototypical example. We show that since stationary solutions of the Fitzhugh-Nagumo model are described by self-adjoint operators, the methods of equilibrium propagation for performing credit assignment can be applied. Furthermore, for Fitzhugh-Nagumo networks with the topology of a deep residual network, we show that the steady state solutions admit a (spatial) Hamiltonian, and thus the methods of Hamiltonian Echo Backpropagation can be applied. We end by deriving an explicit layer-wise Hamiltonian recurrence relation governing inference for stationary solutions of both deep Fitzhugh-Nagumo networks and deep Energy-Based Models.
\end{abstract}

\section{Introduction}

The question of how biological brains coordinate their synaptic updates across the brain in order to perform learning is a major open question in neuroscience, and is also relevant to machine learning. In machine learning, one typically uses stochastic gradient descent (SGD) to perform coordinated updates to all synapses with the goal of minimizing a loss or error function. SGD has the benefit of being both conceptually simple and highly effective in training deep or hierarchical neural networks, and is even able to train spiking and neuromorphic architectures \citep{lee2016training, wunderlich2021event}. 

However, in neuroscience, it is well-known that the backpropagation algorithm, which is used to estimate the gradients that SGD requires to coordinate its weight updates, is not biologically plausible \citep{grossberg1987competitive}. The primary reason for this is that backpropagation is a nonlocal algorithm: it requires an explicit "backward" graph along which to propagate gradient information, which must be matched to the forward graph at all times. This has led some neuroscientists to conclude that SGD itself is not biologically plausible as a learning algorithm. However, various alternatives to backpropagation have been proposed that are able to provide gradient estimates for biologically plausible synaptic updates. A summary of recent work can be found in \citet{lillicrap2020backpropagation}.

\subsection{Related Work}

One promising alternative to backpropagation for training biologically plausible neural networks is Equilibrium Propagation (EqProp) \citep{scellier2017equilibrium}. EqProp applies to the class of networks known as Energy-Based Models (EBMs), such as Hopfield networks and their modern variants \citep{krotov2016dense}. 

Despite its restriction to EBMs, EqProp possesses several properties which make it attractive as a candidate for neural learning. Perhaps most importantly, it is fully local in its weight updates, i.e. the update for a given synapse in the network depends only on the local activity difference of the neurons to which it is connected. EqProp also possesses lower variance in the gradient estimate compared to node and weight perturbation and REINFORCE-like algorithms, which makes it scalable to neural networks of practical relevance \citep{hoier2026training, scellier2023energy, laborieux2021scaling}. At the same time, it yields low-bias gradient estimates compared to algorithms such as Feedback Alignment and Target Propagation \citep{laborieux2021scaling}. Lastly, the same network performs both inference and gradient estimation, so there is no need for a separate backward graph for propagating gradients.

Another approach, with a long history of use in neuroscience, is Predictive Coding \citep{friston2003learning,bogacz2017tutorial,millidge2021predictive}. EqProp and Predictive Coding are closely related, as both can be formulated in terms of a bi-level optimization problem: an inner optimization problem in which an energy function is minimized during inference, and an outer optimization problem where the weights are updated to perform learning. As a result, Predictive Coding and EqProp share many theoretical and empirical properties, such as equivalence to backpropagation under certain conditions \citep{millidge2022backpropagation}, and the fact that both algorithms optimize a well-defined objective function during learning. It is worth noting that Predictive Coding, unlike EqProp, is capable of training feedforward networks.

However, compared to Predictive Coding and many other alternatives to backpropagation, EqProp has the advantage of not requiring a separate, learned backward graph solely for propagating gradient information. The mechanism by which EqProp achieves this is by exploiting the underlying \textit{self-adjointness} of Energy-Based Models. This self-adjointness means these models, differentially, are equal to their own backward graph: their underlying activation Jacobians are symmetric. This is a general property of so-called \textbf{gradient systems}, which follow dynamics determined by the gradient of a scalar function with respect to their state variables. Note that this gradient following is different from gradient descent learning. In gradient descent learning, it is the \textit{synapse} -- i.e., weight --  dynamics which follow the gradient of a loss function. In contrast, in EBMs it is the neurons -- i.e., activations -- which follow the gradient of an energy function to perform inference.

% Expand and summarize efforts to extend equilibrium propagation

\subsection{Gradient Systems and Equilibrium Propagation}

Gradient systems, where the dynamics follow the gradient of an energy function, have the following interesting property: the gradient of an \textit{arbitrary} scalar function defined on a subset of the system's state variables can be propagated backwards, physically through the network, to any other subset of the state variables without requiring a separate backward graph \citep{scellier2017equilibrium, bengio2015early}. Notably, this property enables one to perform local credit assignment in this class of networks, where the function defined on the state variables can be the loss or reward function of a neural network model.\footnote{In other words, it shows that the loss gradient for any node in the network can be accurately calculated using only information that is locally available to that node, without requiring a separate backward graph containing copies of the transposed weights $W_k^T$ and activation function derivatives $f'(h)$.}

This property was originally proved by Scellier and Bengio \citep{scellier2017equilibrium} in energy-based models of neural networks, and was quickly shown to extend to many physical systems of interest such as optical networks \citep{hughes2018training}, resistive electrical circuits \citep{kendall2020training}, Ising machines \citep{laydevant2024training}, mechanical mass-spring oscillators \citep{altman2024experimental}, quantum circuits \citep{wanjura2025quantum, scellier2024quantum}, and other systems of a physical origin \citep{stern2023learning}. 

More recently, Hamiltonian systems were also shown to support an analogous self-adjoint algorithm for propagating gradient information, called Hamiltonian Echo Backpropagation (HEB) \citep{lopez2023self}, which exploits the time-reversibility of a Hamiltonian system to perform credit assignment across time.

These efforts represent significant progress in solving the credit assignment problem in biophysical models of neural networks, and have led to new approaches for the design of neuro-inspired computing architectures  \citep{yi2023activity, martin2021eqspike}. 

From these properties, we believe methods which use the self-adjoint structure of EBMs and Hamiltonian models to perform gradient estimation are promising candidates for explaining how gradient-based learning could originate in simple biological networks. However, \textit{there remains a gap between the classes of systems covered by EqProp and HEB, and real, biophysical models of neural circuits}. Notably, \textbf{there is, to date, no existing method which is able to simultaneously incorporate nonlinear dissipation, gain, and time-varying components} (all present in real biophysical neuronal networks) into a single unified framework capable of performing credit assignment using self-adjoint methods.

\subsection{The Fitzhugh-Nagumo Model}

Unlike gradient systems and conventional Hamiltonian systems, real biophysical neurons tend to simultaneously possess nonlinear dissipation, gain, and energy-storage components such as capacitances. This makes their treatment using techniques like Equilibrium Propagation and Hamiltonian Echo Backpropagation difficult as, in general, their dynamics are not self-adjoint, and therefore cannot be given directly in terms of the gradient of a scalar energy function. 

As a canonical example of this, the Fitzhugh-Nagumo model \citep{fitzhugh1961impulses, nagumo1962active} is the simplest biophysical neuron model capable of generating action potentials. It possesses these three categories of nonlinear dissipation, gain, and energy-storage components, and as a result is capable of displaying a wide array of complex dynamics, including Turing patterns, spirals, traveling waves, solitons (traveling spikes or standing pulse solutions), and critical dynamics \citep{cebrian2024six}. The Fitzhugh-Nagumo equations are given as:

\begin{align*}
    \frac{du}{dt} &= u - u^3 - v \\
    \frac{dv}{dt} &= \varepsilon(u - \alpha v - \beta)
\end{align*}

As one can see, the system of equations defined by the Fitzhugh-Nagumo model is not self-adjoint: its Jacobian is not symmetric with respect to the state variables. However, it is part of a large class of (active) reaction-diffusion systems which possess what has been termed "skew-gradient" structure \citep{yanagida2002standing, yanagida2002mini}. Skew-gradient systems can be described as partitioning the system's state variables into two parts, one describing an "activator" species, and the other describing an "inhibitor" species, with one species following the positive gradient and the other following the negative gradient of an energy function. For example, in the Fitzhugh-Nagumo model, the activator corresponds to the membrane potential of the neuron and the inhibitor corresponds to the recovery variable of the neuron. Instead of being gradient systems, where all state variables follow the gradient of an energy function, they are mini-maximizers with respect to the energy function \citep{yanagida2002mini}. Some variables (the activators) seek to maximize the energy, while others (the inhibitors) seek to minimize the energy.

In this work, we extend Equilibrium Propagation and Hamiltonian Echo Backpropagation to the stationary (steady-state) solutions of such skew-gradient systems, and analyze and train a skew-gradient based deep neural network based on the Fitzhugh-Nagumo model using Equilibrium Propagation. By exploiting the underlying spatial Hamiltonian structure of the stationary states, we then derive a forward layer-wise recursion which is able to perform inference in a single forward pass, given appropriate initial conditions, without relying on iterative temporal convergence to a fixed point, as is typical in the inference process of energy-based models. Finally, we apply these ideas to the more conventional energy-based models of \citet{scellier2017equilibrium, bengio2015early}.

\section{Main Result}

A network of Fitzhugh-Nagumo neurons with diffusive (resistive) coupling can be written in the following form, as described in Yanagida \citep{yanagida2002standing, yanagida2002mini}.

\begin{equation}
    T\mathrm{u}_t=D\Delta\mathrm{u} + \mathrm{f(u)}
\end{equation}
where $T$ and $D$ are positive, diagonal matrices, $\Delta$ is a Laplacian operator, which describes the neuron-to-neuron coupling, and $f(u)$ is a nonlinear function which can be expressed as the gradient of a scalar function $F$, usually called the \textit{free energy}, multiplied by a matrix $Q$:

\begin{equation}
    \mathrm{f(u)}=Q\frac{\partial F}{\partial \mathrm{u}},
\end{equation} with $Q$ satisfying $Q^2=\mathbb{I}_2$. For example, in our case, we choose $\mathrm{u}=\begin{bmatrix} u \\ v \end{bmatrix}$, and $Q=\begin{bmatrix} 1 & 0 \\ 0 & -1\end{bmatrix}$. For notational convenience, we will refer to the upright $\mathrm{u}$ as the combined set of activator and inhibitor concentrations, and the script $u$ and $v$ for the activator and inhibitor concentrations, respectively.

Thus, the Fitzhugh-Nagumo model with a single activator species $u$ and a single inhibitor species $v$ is given as follows:

\begin{equation*}
    \begin{bmatrix}
        \tau_1 & 0 \\ 0 & \tau_2
    \end{bmatrix}
    \begin{bmatrix}
        u_t \\ v_t
    \end{bmatrix}
    =
    \begin{bmatrix}
        d_1 & 0 \\ 0 & d_2
    \end{bmatrix}
    \begin{bmatrix}
        \Delta u \\ \Delta v
    \end{bmatrix}
    +
    \begin{bmatrix}
        1 & 0 \\ 0 & -1
    \end{bmatrix}
    \begin{bmatrix}
        \nabla_u F \\ \nabla_v F
    \end{bmatrix}
\end{equation*}

At a conceptual level, we will take the perspective of Nagumo and physically interpret the Fitzhugh-Nagumo model as an electrical circuit with a tunnel diode nonlinearity and resistive coupling. In the electrical circuit interpretation, we can think of the activator and inhibitor concentrations $u$ and $v$ as the node voltages (or loop currents) of a circuit, as shown in Fig. 1B. 

We will use the discrete Laplacian \citep{muolo2024turing} in place of the usual continuous Laplacian, since we are primarily interested in discrete networks. We can construct the Laplacian of a generic weighted, undirected (diffusive) graph using the incidence matrix of the graph, and explicitly write out the full set of discrete Fitzhugh-Nagumo equations in terms of the graph Laplacian of the network.

\begin{equation*}
    \begin{bmatrix}
        \tau_1 u_t \\ \tau_2 v_t
    \end{bmatrix}
    =
    \begin{bmatrix}
        d_1 \mathbb{I}_n & 0 \\ 0 & d_2 \mathbb{I}_n
    \end{bmatrix}
    \begin{bmatrix}
        L_1 u \\ L_2 v
    \end{bmatrix}
    +
    \begin{bmatrix}
        \mathbb{I}_n & 0 \\ 0 & -\mathbb{I}_n
    \end{bmatrix}
    \begin{bmatrix}
        \nabla_u F \\ \nabla_v F
    \end{bmatrix}
\end{equation*}

Here, $\mathbb{I}_n$ is the $n\times n$ identity matrix, and $u$ and $v$ are $n$-dimensional vectors for a network with $n$ nodes. The Laplacians $L_1$ and $L_2$ govern the network topology for the activator and inhibitor, respectively, and can be written in terms of the corresponding branch-node incidence matrix $B_i$ and a diagonal matrix of positive conductances (weights) of the edges $Y_i$, where $i=1$ corresponds to the activator and  $i=2$ the inhibitor.

\begin{figure}[!t]
    \centering
    \includegraphics[width=1.0\linewidth]{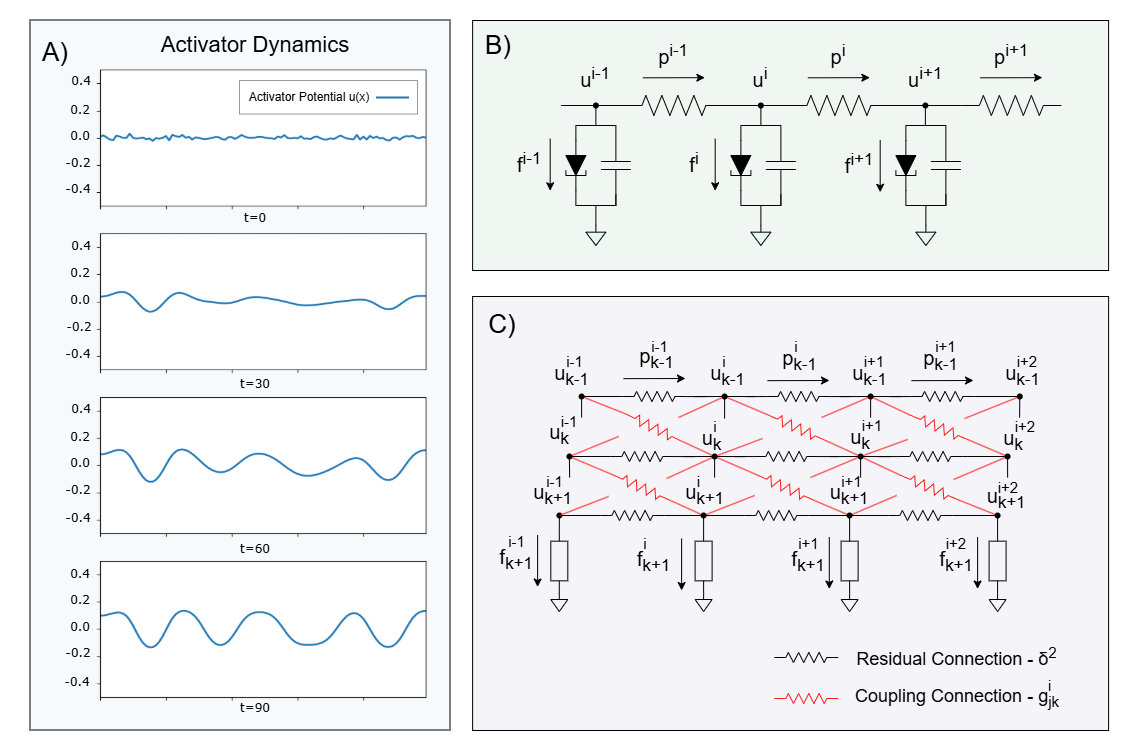}
    \caption{A) Time dynamics (time increasing downwards) of the Fitzhugh-Nagumo model defined on a 1-dimensional spatial interval. The parameter settings of the model are tuned to display convergence to a stationary 1-dimensional Turing pattern, i.e. spatial oscillations. B) Discrete 1D spatial model of coupled Fitzhugh-Nagumo neurons (activator only shown), corresponding to a 1-dimensional path graph. Activator variable $u$ at position $i$ in the path is denoted $u^i$. Inhibitor $v^i$ is not explicitly shown, and takes the same form, but with no nonlinearity. The nonlinear current through the tunnel diode is denoted by $f^i$. The momentum variable $p^i$ is the voltage drop between two adjacent nodes $u^{i+1} - u^i$ along the path. C) Coupled path graphs with coupling connections denoted by $g^i_{jk}$, connecting node $j$ of layer $i$ to node $k$ of layer $i+1$, forming a deep residual network of FHN neurons.}
    \label{fig:placeholder}
\end{figure}

\begin{equation*}
    L_i = B_i^TY_iB_i
\end{equation*}

As a consequence of this form, our Laplacian matrices are real-symmetric. Now, we can add an input current to the network. We will represent this input current as a vector of the same dimension as the activator $u$, and therefore we can inject current into any or all of the $u$ nodes in the network, whether they be input, hidden, or output nodes. While we restrict our input current to the activator variables $u$ for simplicity, the inhibitor variables $v$ can be treated in the same way. Our system is now given by:

\begin{equation*}
    T\mathrm{u}_t=D L \mathrm{u} + \mathrm{f(u)} - I,
\end{equation*} 

where again, $\mathrm{u}$ is defined as $[u,v]^T$, $L$ is a block diagonal matrix containing $L_1$ and $L_2$, and $I=[I_u, 0]^T$ is the input current, assumed constant in time. Note that $d_1$ and $d_2$ in $D$ are global scaling factors for the conductances (diffusion terms) for $u$ and $v$, multiplying their respective Laplacians $L_1$ and $L_2$. Throughout the rest of this section, we restrict our attention to stationary solutions of (6), given by:

\begin{equation*}
    DL\mathrm{u_s} + \mathrm{f(u_s)}=I
\end{equation*}

Here, $\mathrm{u_s}$ is a stationary solution of (6). We can analyze how small changes in our (steady-state) input current $I$ affect the steady state solution by linearizing (7) around $u_s$.

\begin{equation*}
    \left( DL + \frac{\partial\mathrm{f(u_s)}}{\partial\mathrm{u}} \right)\delta \mathrm{u} = \delta I
\end{equation*}

We obtain:

\begin{equation*}
    \begin{bmatrix}
        d_1L_1 + f'(u) & \mathbb{I} \\ -\mathbb{I} & d_2L_2 - \gamma
    \end{bmatrix}
    \begin{bmatrix}
        \delta u \\ \delta v
    \end{bmatrix}
    =
    \begin{bmatrix}
        \delta I_u \\ 0
    \end{bmatrix},
\end{equation*}

where $f'$ and $\gamma$ are given by the derivative of $\mathrm{f(u)}$ with respect to $u$ and $v$, respectively. Note that, as a whole, the full linearized system matrix in (9) is not symmetric. This means that it is not a gradient system, and is therefore not governed by a Lyapunov function whose value strictly decreases with time. Instead, it is a \textit{skew-gradient} system, where one set of variables aims to minimize, and the other set of variables aims to maximize, the free energy function $F$ \citep{yanagida2002mini}.

We can solve (9) for the activator variable $u$, yielding the activator's linearized response to the current perturbation at the steady state $\mathrm{u_s}$. In particular, we find:

\begin{equation*}
    \delta u = \left( \left( d_1L_1 + f'(u_s)\right) + \left( d_2L_2 - \gamma \right)^{-1} \right)^{-1}\delta I_u = M^{-1}\delta I_u
\end{equation*}

Since $L_1$ and $L_2$ are symmetric by construction, and $f'(u)$ and $\gamma$ are diagonal matrices, and since the inverse of a symmetric matrix is again symmetric, the matrix $M^{-1}$ must therefore be symmetric. Note that $M^{-1}$ is a \textit{submatrix} of the full system response matrix, corresponding to current injection of the activator species $I_u$ and subsequent change in the steady state of the activator $u_s$.

This result tells us that the network's \textit{effective response matrix} for the activator variables (the Jacobian of the activator variables $u_s$ with respect to the injected current $I_u$) is self-adjoint for all nodes, and across all settings of the Fitzhugh-Nagumo parameter values, as long as the network is at a steady state. Thus, the methods of Equilibrium Propagation can be applied. A summary of why EqProp-like learning rules follow directly from the self-adjointness of the underlying network is now given.

Let us assume we wish to compute the gradient of a loss function $\mathcal{L}$ with respect to a parameter at layer $i$, say $g_i^k$. The loss $\mathcal{L}$ is a function of the network's output variables. Let the output variables be denoted $u_y$. These output variables are in turn a function of the network's hidden state variables at layer $i$, let these be denoted $u_{h_i}$. Finally, the hidden state at layer $i$ is a function of the parameter value $g_i^k$. Then we have

\begin{equation*}
    d\mathcal{L} = \frac{\partial \mathcal{L}}{\partial u_y} \frac{\partial u_y}{\partial u_{h_i}} \frac{\partial u_{h_i}}{\partial g_i^k} dg_i^k
\end{equation*}

This equation describes the differential change in the value of the loss $\mathcal{L}$ via a change in the parameter of interest, $g_i^k$. In other words, this is the forward directional derivative of the loss with respect to a change in the parameter of interest. Since there are many different parameters in the network, and only a single scalar loss function, the adjoint of this equation is typically used, which is also known as backward-mode differentiation. This is given by transposing the above relation to yield:

\begin{equation}
    \frac{\partial \mathcal{L}}{\partial g_i^k} = \left(\frac{\partial u_{h_i}}{\partial g_i^k}\right) ^T  \left(\frac{\partial u_y}{\partial u_{h_i}}\right)^T \left(\frac{\partial \mathcal{L}}{\partial u_y}\right)^T
\end{equation}

We see that in backward mode differentiation, the network's forward Jacobian $\frac{\partial u_y}{\partial u_{h_i}}$ is replaced by its transpose. For a network to be self-adjoint, this matrix must be symmetric, and therefore the transposed quantity is identical to the network's forward Jacobian. This means that we can perform \textit{a single inference-time perturbation} to the output neurons, proportional to $\frac{\partial \mathcal{L}}{\partial u_y}$, which travels backwards through $\frac{\partial u_y}{\partial u_{h_i}} = \left(\frac{\partial u_y}{\partial u_{h_i}}\right)^T$, carrying gradient information to the rest of the nodes in the network. These perturbations can be locally measured at each hidden layer and used to assign credit to the parameters via Eq. (3). The general mechanism of encoding gradient information in terms of perturbations or activity differences is known as NGRAD (Neural Gradient Representation by Activity Differences), and is discussed in general in \citet{lillicrap2020backpropagation}.

Because there is only a single perturbation required for the entire gradient estimation step, EqProp and related methods have much lower variance than node or weight perturbation methods, where the variance in the gradient estimate scales with the number of parameters or nodes in the network \citep{lillicrap2020backpropagation}.

\section{Training Deep Fitzhugh-Nagumo Networks}

In order to demonstrate that the above theoretical analysis holds experimentally, we demonstrate the training of deep networks of Fitzhugh-Nagumo neurons, in an analogous fashion to deep EBMs, using Equilibrium Propagation. We mirror the experiments done in \citet{scellier2017equilibrium} by training a deep Fitzhugh-Nagumo network with 5 hidden layers on MNIST using EqProp. Code is available on Github: \href{https://github.com/jackdkendall/Self-Adjoint-Learning}{Self-Adjoint-Learning}

For the deep FHN networks we train, we initialize the FHN parameter values $\delta$, $\varepsilon$, $\alpha$, $\beta$ in the Turing pattern forming regime. We keep these parameters fixed during training. We initialize the weights normally around zero, as in a standard EBM. Since the skew-gradient formulation of the FHN nonlinearity already admits negative conductances, we do not clamp parameter conductances to be strictly positive, and we simply treat negative conductances as effective gain elements. For Equilibrium Propagation, we find that a large value of $\beta_{nudge}$, near 0.9, works best. We use the centered difference formulation of EqProp, and consistent with prior work on EqProp training, ensure that earlier layers have larger learning rates than later layers.

\begin{table}[ht!]
\begin{center}
\begin{tabular}{||c c c||} 
 \hline
 Architecture: & 784-512-512-512-512-512-10 &  \\ [0.5ex] 
 \hline
 layerwise lr's: & [1e-2, 1e-3, 2e-4, 1e-4, 5e-5] &  \\
 \hline
 $\delta$, $\varepsilon$, $\alpha$, $\beta$: & 0.75, 0.85, 1.08, 0.0 &  \\
 \hline
 $\beta_{nudge}$, num iters, nudge iters: & 0.9, 55, 14 &  \\
 \hline
 \textbf{MNIST Test Error:} & \textbf{2.8\%} $\pm$ (0.2) &  \\ 
 \hline
\end{tabular}
\end{center}
\caption{Optimal hyperparameters and MNIST test error for a 5-layer Fitzhugh-Nagumo Network trained with Equilibrium Propagation}
\end{table}

\vspace{3mm}

For inference, we follow the time-dynamics defined by the FHN model to approximate convergence, which typically takes ~50 iterations, and we use a fixed step size of 0.1. Our weight initialization scale (scalar multiplier of standard normal initialization) was chosen to be ~0.01, with 0.014 showing the best results. We noticed in some instances, the occurrence of loss spikes during training. We speculate that these may be caused by the FHN model leaving the steady-state dynamical regime it was initialized in, via accumulation of weight updates which increase the effective gain of the network. More experiments are needed to investigate this behavior.

\section{Hamiltonian Inference}

While Energy-Based Models have a number of desirable properties, such as good sample diversity in generative models and natural compositionality properties \citep{du2020compositional}, they have not seen widespread use in industry, because of their expensive inference process relative to feedforward networks. Whereas feedforward networks require a single forward pass through the network to perform inference, EBMs require many forward passes to reach convergence to the energy minimum. 

We will show that deep EBMs possess a feedforward Hamiltonian formulation, \textbf{which is equivalent to the original EBM up to specification of boundary conditions}. This Hamiltonian formulation is entirely feedforward in the inference process, but requires the specification of an additional "momentum" variable at the input to the network, as it turns the original EBM's two-point boundary problem into a purely initial value problem. To begin with, we will generalize the Hamiltonian formulation of the FHN from the continuous setting, where it is well established \citep{parra2025spatial}, to the discrete network setting. Then, we will transfer these results to the Energy-Based Models considered by Scellier and Bengio. 

\subsection{Hamiltonian Formulation of the Fitzhugh-Nagumo Model}

The fact that the Fitzhugh-Nagumo model admits a Hamiltonian formulation was first noticed by Yanagida \citep{yanagida2002standing, yanagida2002mini}. There is now a large body of literature which characterizes the stationary solutions of the Fitzhugh-Nagumo model in terms of \textit{spatial dynamics}, i.e. they describe the spatial evolution of the solutions $\mathrm{u_s}(x)$ in terms of a Hamiltonian which is conserved across a spatial dimension $x$, where $x$ now plays the role of "time" in a typical nonlinear Hamiltonian problem \citep{kuwamura2004turing, karasozen2017structure, parra2025spatial}. However, these results hold for continuous 1D systems, and to the author's knowledge, there have been no attempts at generalizing this framework to the setting of discrete graphs. So, we will now apply the Hamiltonian formalism of the Fitzhugh-Nagumo model to the discrete undirected graph Laplacian setting. We will analyze the autonomous case, where $I=0$.

In \citep{kuwamura2004turing}, the 1-dimensional continuous Laplacian is considered, with the spatial coordinate given by $x$. Steady-state solutions of the continuous 1-dimensional Fitzhugh-Nagumo model are described by:

\begin{equation*}
    D \frac{\partial^2}{\partial x^2} \mathrm{u} = -\mathrm{f(u)}
\end{equation*}

where $D$ and $\mathrm{f(u)}$ obey the constraints given in (1). For our analysis, we will replace the 1-dimensional continuous Laplacian $\frac{\partial^2}{\partial x^2}$ with a generic weighted, undirected graph Laplacian $L$. We will assume the Laplacian has been grounded, or reduced, such that its zero eigenvalue is removed. Since $L$ is positive definite, it possesses a unique non-negative square root, which we will denote $L^{1/2}$, such that $L^{1/2}(L^{1/2})^T = L^{1/2}L^{1/2} = L$. We will define the "spatial velocity" $e$ to be the square root of the Laplacian acting on $\mathrm{u}$. This will be our discrete analog of $\mathrm{u}_x$, the spatial gradient of $\mathrm{u}$ along $x$.

\begin{equation*}
    \mathrm{e} := L^{1/2}\mathrm{u}
\end{equation*}

Now, following \citet{kuwamura2004turing}, we define a new variable $Z$ as:

\begin{equation*}
    Z = \begin{bmatrix}
        \mathrm{u} \\ \mathrm{e}
    \end{bmatrix}
\end{equation*}

We then define the spatial gradient of $Z$ as:

\begin{equation*}
    Z_x = 
    \begin{bmatrix}
        L^{1/2}\mathrm{u} \\ L^{1/2}\mathrm{e}
    \end{bmatrix}
    =
    \begin{bmatrix}
        \mathrm{e} \\ L\mathrm{u}
    \end{bmatrix}
\end{equation*}

Using this notation, we can break the set of n second-order difference equations into a set of 2n first-order difference equations:

\begin{align*}
    \mathrm{u}_x &= \mathrm{e} = L^{1/2}\mathrm{u} \\
    D\mathrm{e}_x &= DL^{1/2}\mathrm{e} = -\mathrm{f(u)}
\end{align*}

Indeed, we see that $D L^{1/2}(L^{1/2}\mathrm{u}) = DL\mathrm{u}$. Next, define the following (spatial) Hamiltonian:

\begin{equation*}
    H := \frac{1}{2} \mathrm{e}^TDQe + F(\mathrm{u})
\end{equation*}

Where

\begin{equation*}
    F := \int_0^{\mathrm{u}} \mathrm{f(u')} d \mathrm{u'} = \frac{1}{2} \alpha u^2 - \frac{1}{4} \alpha u^4 - uv + \frac{1}{2} \beta v^2
\end{equation*}

This corresponds to the Fitzhugh-Nagumo model Free Energy function given in \citet{yanagida2002standing}. Finally, we can write the Fitzhugh-Nagumo equations in their Hamiltonian, or symplectic, form:

\begin{equation*}
    \begin{bmatrix}
        0 & QD \\
        -QD & 0
    \end{bmatrix}
    \begin{bmatrix}
        \mathrm{u}_x \\ \mathrm{e}_x
    \end{bmatrix}
    =
    -\begin{bmatrix}
        \frac{\partial H}{\partial \mathrm{u}} \\ 
        \frac{\partial H}{\partial \mathrm{e}}
    \end{bmatrix}
    =
    -\frac{\partial H}{\partial Z}
\end{equation*}

Or, more compactly, as:

\begin{equation}
    KZ_x = -\frac{\partial H}{\partial Z}
\end{equation}

The existence of a Hamiltonian formulation in the coordinates $Z$ has several important implications for discrete 1-dimensional Laplacians. First, it means if one knows the pair $(\mathrm{u,e})$ at any node $i$ in the path graph, the values of $(\mathrm{u,e})$ on the entire graph can be constructed by recursively solving Eq. (4) for $Z_x$, starting from $i$, and integrating $(u+u_x,e+e_x)$ along the graph, rather than iteratively simulating the dynamics of the entire network to convergence, as is commonly done in inference of energy-based models. Second, the spatial Hamiltonian $H$ is constant (for stationary solutions) along the path graph. In other words, the sum of all currents into each node of the graph is equal to zero. Third, the spatial evolution of $(\mathrm{u,e})$ along the graph is formally equivalent to a conservative nonlinear oscillator problem in discrete time, for Turing pattern steady-state solutions. 

We will now use these ideas to develop a Hamiltonian approach for an example of a Fitzhugh-Nagumo network with the structure of a deep residual neural network. We will start with a discrete 1-dimensional (path graph) Laplacian $L$ with $n$ nodes given by\footnote{Note that we treat the boundary of the Laplacian as if it were "impedance matched", i.e. a ghost node is added to the boundary nodes which double the value of the Laplacian's diagonal terms at the first and last row, making the diagonal constant.}:

\begin{equation*}
    L  = \begin{bmatrix}
    2 & -1 & 0 & 0 & 0 & 0\\
    -1 & 2 & -1 & 0 & 0 & 0\\
    0 & -1 & 2 & \ddots  & 0 & 0\\
    0 & 0 & \ddots  & \ddots  & -1 & 0\\
    0 & 0 & 0 & -1 & 2 & -1\\
    0 & 0 & 0 & 0 & -1 & 2
    \end{bmatrix}
\end{equation*}

We will consider the result of coupling many of these 1D path graphs together in parallel. To simplify the analysis in this section, we will use the following formulation of the Fitzhugh-Nagumo model, used in \citet{parra2025spatial} to study the spatial dynamics of the steady states.
\begin{align*}
    \delta^2\Delta u + u - u^3 - v &= 0 \\
    \Delta v+ \varepsilon(u - \alpha v - \beta) &= 0
\end{align*}

First, we can extend our 1-dimensional network by expanding our path graph Laplacian to include $m$ separate path graphs, initially uncoupled from one another, each with path length $n$. Then, we can introduce layer-wise coupling between adjacent sets of nodes in the independent path graphs, i.e. by creating a Laplacian with the structure of a deep residual neural network, as shown in Fig. 1C. For stationary solutions corresponding to Turing patterns, the resulting system is formally equivalent to a \textit{coupled} system of nonlinear conservative oscillators in discrete time.

We will now derive a layer-wise recursion for steady-state solutions of this network. This means that instead of simulating the (skew-gradient) time dynamics of the network until convergence, which requires many forward passes over the network until the steady state is found, we can directly compute the steady state solution of each layer directly from the steady state of the previous layer. That is, we can perform inference in a single forward pass, given an appropriate initial condition, which we will define as the initial position and momentum $(u^0, p^0)$. Thus, instead of defining our network only in terms of the $\mathrm{u}^i$ variables, we must describe it in terms of the phase-space variables $(\mathrm{u}^i, \mathrm{p}^i)$.

Next, since the network is at a steady state, we note that the total current through the capacitors is equal to zero. We define the value of the activator at node $i$ along path $k$ to be $u_k^i$. The current through the nonlinear tunnel diode branch $f^i_k$ is equal to the gradient of the free energy $F$ with respect to $u_k^i$. 

The momentum variable, we define to be the voltage drop between adjacent nodes in the same path, $p^i := u^{i+1} - u^i$. Here, $u^i$ and $p^i$ are vectors of node activator variables and edge drops, respectively, of all nodes at layer $i$ in their path graph. The coupling matrices couple the nodes at layer $i$ to the nodes at layer $i+1$, and are given by $g^i_{jk}$. 

Now, the condition of the spatial Hamiltonian being constant along the path in the abstract Laplacian analysis given in Section 3.1 is equivalent to the statement that for all nodes in the network, the currents into that node sum to zero. Or in other words, it is a consequence of Kirchhoff's current law. This can also be viewed as a homology relation on the network \citep{smale1971mathematical}.

For uncoupled path graphs, one path of which is shown in Fig. 1B, the total current at node $i+1$ in path $k$ of the graph is given by:

\begin{equation*}
    \delta^2p^{i+1}_k - \delta^2p^i_k - f^{i+1}_k = 0
\end{equation*}

From this relation, the voltage drop to the next node in the path graph, $p^{i+1}_k$, can be computed from $f^{i+1}_k$ and $p^i_k$. 

For the case of coupled paths, the total current is more complicated, but illuminates the connection between the iterative energy-based dynamics and the direct Hamiltonian computation of the steady-state solution. The total current at node $i+1$ for coupled paths is given by:

\begin{dmath}
    \delta^2p^{i+1}_k - \delta^2p^i_k - f^{i+1}_k + \sum_{j}^m \left( u_j^i - u_k^{i+1} \right) g_{kj}^i + \sum_{j} \left(  u^{i+2}_j - u^{i+1}_k \right) g_{jk}^{i+1}=0
\end{dmath}

Using the definitions of $p^i_k$ and $p^{i+1}_k$, we can rewrite the two sums above as:

\begin{equation*}
    \sum_{j}^m \left( u_j^i - u_k^{i+1} \right) g_{kj}^i = \sum_{j}^m \left( (u^{i+1}_j - p^{i}_j) - u_k^{i+1} \right) g_{kj}^i
\end{equation*}

\begin{equation*}
    \sum_{j} \left(  u^{i+2}_j - u^{i+1}_k \right) g_{jk}^{i+1} = \sum_{j} \left( (u^{i+1}_j + p^{i+1}_j) - u^{i+1}_k \right) g_{jk}^{i+1}
\end{equation*}

This reduces Eq. (5) to the following equation, which is only a function of $p^i$, $p^{i+1}$, and $u^{i+1}$, and where $(G^i)_{jk} := g_{jk}^i$, $\hat{g}^i$ is the sum along the columns of $G^i$, and $\tilde{g}^{i}$ is the sum along the rows of $G^{i}$.

\begin{dmath*}
    \left( G^{i+1} + \delta^2\mathbb{I}_m \right) p^{i+1} = \left( (G^i)^T + \delta^2 \mathbb{I}_m \right) p^i + f^{i+1} - \left( (G^i)^T + G^{i+1} - diag(\hat{g}^i + \tilde{g}^{i+1}) \right) u^{i+1}
\end{dmath*}

Let $M^i = (G^{i+1} + \delta^2\mathbb{I}_m)^{-1}$, $N^i = ((G^i)^T + \delta^2\mathbb{I}_m)$, and $O^i = ((G^i)^T + G^{i+1} - diag(\hat{g}^i + \tilde{g}^{i+1}))$. Then we can solve for $p^{i+1}$ as

\begin{equation*}
    p^{i+1} = M^i N^ip^i + M^i f^{i+1} - M^i O^i u^{i+1}
\end{equation*}

Given a $u_i$ and a $p_i$ at any layer $i$, we can then compute the steady-state solution of the network at layer $i+1$ via the following layer-wise recursion relation. This yields the final Hamiltonian recurrence relations for the deep residual Fitzhugh-Nagumo network for the activator variables:
\begin{align*}
    u^{i+1} &= u^i + p^i \\
    p^{i+1} &= M^i N^ip^i + M^i f^{i+1} - M^i O^i u^{i+1}
\end{align*}

And for the inhibitor variables, which can be treated similarly, but with no inter-path coupling:

\begin{align*}
    v^{i+1} &= v^i + q^i \\
    q^{i+1} &= \varepsilon (u^{i+1} - \alpha v^{i+1} - \beta)
\end{align*}

\subsection{Hamiltonian Formulation of Deep EBMs}

Consider the energy function defined in \citet{scellier2017equilibrium}, which is a kind of modified Hopfield energy, originally used in \citet{bengio2015early}.

\begin{equation*}
    E = \frac{1}{2}\sum_i u_i^2 - \frac{1}{2}\sum_{i\neq j}W_{ij}\rho(u_i)\rho(u_j) - \sum_ib_i\rho(u_i)
\end{equation*}

The gradient of the energy function with respect to $u_i$, which drives the dynamics of the network, is then given by:

\begin{equation*}
    \frac{\partial E}{\partial u_i} = u_i - \rho'(u_i) \left( \sum_j W_{ij} \rho(u_j) - b_i \right)
\end{equation*}

We will now show how to perform the above Hamiltonian analysis for this energy-based model. First, we will perform a change of variables which will allow us to transfer the nonlinearity from the interaction term involving $W_{ij}$ to the local term involving $u_i$. Assuming the nonlinear activation function $\rho(u_i)$ is strictly monotonic (e.g. sigmoidal), we can make the following change of variables:
\begin{equation*}
    v_i := \rho(u_i) \implies  u_i = \rho^{-1}(v_i)
\end{equation*}

The gradient of the energy function with respect to $u_i$ can be written in terms of $v_i$ using the chain rule as:

\begin{equation*}
    \frac{\partial E}{\partial u_i} = \frac{\partial E}{\partial v_i} \frac{\partial v_i}{\partial u_i} = \frac{\partial E}{\partial v_i} \rho'(u_i)
\end{equation*}

Substituting this into the LHS of (43) and the definition of $v_i$ into the RHS of (43):

\begin{equation*}
    \frac{\partial E}{\partial v_i} \rho'(u_i) = \rho^{-1}(v_i) - \rho'(u_i) \left( \sum_j W_{ij} v_j - b_i \right)
\end{equation*}

\begin{equation*}
    \frac{\partial E}{\partial v_i} = \frac{\rho^{-1}(v_i)}{\rho'(\rho^{-1}(v_i))} - \sum_jW_{ij}v_j - b_i
\end{equation*}

We can absorb the nonlinear term $\frac{\rho^{-1}(v_i)}{\rho'(\rho^{-1}(v_i))}$ and the bias $b_i$ into a single local nonlinearity $f_i$ at $v_i$:

\begin{equation*}
    \frac{\partial E}{\partial v_i} = f_i(v_i) - \sum_jW_{ij}v_j
\end{equation*}

We will now make the assumption that the network is structured as a \textit{deep} energy-based model, that is, that the weights and activations can be partitioned into layers, and we will assume a constant layer width. The gradient of the energy with respect to the $i$th neuron at layer $l+1$ in the network is then given by:

\begin{equation*}
    \frac{\partial E}{\partial v_i^{l+1}} = f_i(v_i^{l+1}) - \sum_j W_{ji}^l v_j^l - \sum_j W_{ij}^{l+1}v_j^{l+2}
\end{equation*}

In layerwise vector form, this can be written as:

\begin{equation*}
    \frac{\partial E}{\partial v^{l+1}} = f_i(v^{l+1}) - (W^l)^Tv^l - W^{l+1}v^{l+2}
\end{equation*}

As in Section 3.2, we will make the following definition:

\begin{equation*}
    p^l := v^{l+1} - v^l \implies v^{l+1} = v^l + p^l
\end{equation*}

This will allow us to express the terms $v^l$ and $v^{l+2}$ in terms of $v^{l+1}$, $p^l$, and $p^{l+1}$:

\begin{equation*}
    \frac{\partial E}{\partial v^{l+1}} = f(v^{l+1}) - (W^l)^T (v^{l+1} - p^l) - W^{l+1} (v^{l+1} + p^{l+1}) 
\end{equation*}

For stationary solutions of the network, the above energy gradient sums to zero. Rearranging terms, we get:

\begin{equation*}
    0 = f(v^{l+1}) - W^{l+1}p^{l+1} + (W^l)^Tp^l - ((W^l)^T + W^{l+1})v^{l+1}
\end{equation*}

\begin{equation*}
    W^{l+1} p^{l+1} = f(v^{l+1}) + (W^l)^T p^l - ((W^l)^T + W^{l+1})v^{l+1}
\end{equation*}

This allows us to directly obtain the following Hamiltonian recurrence relations on $v^l$ and $p^l$, which gives the equation for exact layer-wise inference of steady-state solutions in deep energy-based models:

\begin{align*}
    v^{l+1} &= v^l + p^l \\
    p^{l+1} &= M^{l} f(v^{l+1}) + N^{l} p^l - (N^l + \mathbb{I}) v^{l+1}
\end{align*}

Where

\begin{align*}
    M^{l} &= (W^{l+1})^{-1} \\
    N^{l} &= (W^{l+1})^{-1}(W^l)^T \\
\end{align*}

We remark that the analogy here is that for energy-based models at their steady-state solution, the state variables are stationary with respect to time. As a result, the gradients of the energy with respect to each of the state variables (which can be interpreted as "currents", if the state variables are interpreted as "voltages"), satisfy a local conservation law: the components of the gradient sum to zero for each node in the network. These additional conservation equations on the gradients can then be used to calculate the components of the gradient along the forward integration direction. Given that the state variable and the momentum variable at one layer are known, the next layer's activation can be calculated from this component of the gradient.

\section{Hamiltonian Simulations}

In this section, we focus on simulations of the Hamiltonian inference process for the deep Fitzhugh-Nagumo networks described above. For the Hamiltonian formulation of Deep EBMs, we defer this discussion to a separate paper which focuses on this topic.

\begin{figure}[!t]
    \centering
    \includegraphics[width=0.8\linewidth]{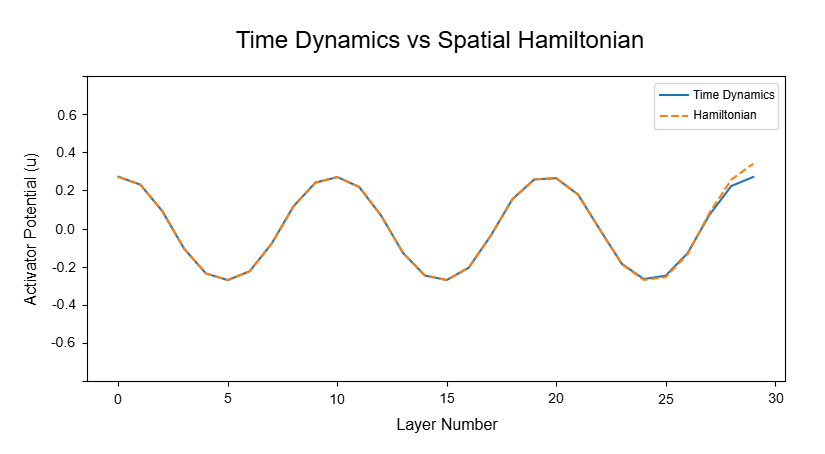}
    \caption{Comparison between the result of a time-dynamics simulation and a Hamiltonian spatial integration in depth for a deep Fitzhugh-Nagumo residual network initialized in the Turing pattern regime. A single neuron's activator potentials are plotted with respect to depth. The network simulated was 30 layers deep and 64 neurons in width. $\sim$ 30 layers was the maximum forward integration depth, beyond which the forward integration diverges. We note that instability of the forward dynamics of deep EBMs has been observed during training of very deep models \cite{innocenti2026mu}.}
    \label{fig:placeholder}
\end{figure}

To show that the Hamiltonian spatial integration of the FHN network from the initial conditions $(u_0,p_0)$ and $(v_0, q_0)$ indeed computes the same solution as the iterative time-based simulation of the steady state, we first perform a time-based simulation of the above equations to calculate the steady-state solution. We then take the first layer's node values $(u_0, v_0)$ along with its residual branch values $(p_0, q_0)$, and use those as the input to the following forward recursion:
\begin{align*}
    u^{i+1} &= u^i + p^i \\
    v^{i+1} &= v^i + q^i
\end{align*}
\begin{align*}
    p^{i+1} &= M^i N^ip^i + M^i f^{i+1} - M^i O^i u^{i+1} \\
    q^{i+1} &= \varepsilon (u^{i+1} - \alpha v^{i+1} - \beta)
\end{align*}

To maintain the correct boundary condition at the initial input (where we have a ghost node), we must modify the first step in the recursion by multiplying by a factor of $1/2$:
\begin{align*}
    p^{1} &= \frac{1}{2}\left[M^0 N^0p^0 + M^0 f^{1} - M^0 O^0 u^{1} \right]\\
    q^{1} &= \frac{1}{2}\left[ \varepsilon (u^{1} - \alpha v^{1} - \beta) \right]
\end{align*}

The results of these simulations are shown in Fig. 2. Note that for networks with a depth of under approximately 30 layers, the spatial integration stays close to the time-based steady-state. However, for deeper networks, the solution eventually diverges, because the forward spatial integration is along an unstable manifold for the Fitzhugh-Nagumo model. More advanced spatial integration schemes or modifying the network architecture, for example to include layer-wise normalization, may alleviate this issue.

We note that the above analysis is performed simply to show that the time-based and layer-wise simulations are equivalent. More practical methods for performing inference directly in the Hamiltonian framework are available, e.g. in \citet{pourcel2025learning}. A discussion of the Hamiltonian approach to inference in deep Energy-Based Models will be discussed in a separate work.

A link to the code used for these simulations can be found at \href{https://github.com/jackdkendall/Self-Adjoint-Learning}{Self-Adjoint-Learning}.

\section{Conclusion}

We have shown that skew-gradient systems, such as the Fitzhugh-Nagumo model, are compatible with the Equilibrium Propagation algorithm at their steady state solutions. Additionally, using techniques from the study of skew-gradient systems, we have also shown how, in deep Energy-Based Models, one can derive a Hamiltonian recurrence relation which allows for exact layer-wise computation of the steady-state solution, provided an additional initial condition on the momentum is given. Thus, the Energy-Based and Hamiltonian formulations differ only via the definition of the boundary conditions for the system: the Energy-Based formulation is a two-point boundary value problem, and the Hamiltonian formulation is a purely initial-value problem. Having an explicit bridge between the Energy-Based (spatial Lagrangian) and Hamiltonian formulations of a network opens the door to applying many useful methods of analysis which rely on a Hamiltonian form to deep Energy-Based Models. Most notably, this includes bifurcation analysis of the uniform network's spatial orbits. We believe this work will provide a useful new tool for training deep networks of activator-inhibitor neurons, and give additional insight into the widespread occurrence of activator-inhibitor structure in pattern formation and biological networks.

\bibliographystyle{tmlr}
\bibliography{references.bib}

\end{document}